\documentclass[11pt,twocolumn,letterpaper]{article}

\usepackage{cvpr}
\usepackage{times}
\usepackage{epsfig}
\usepackage{graphicx}
\usepackage{amsmath}
\usepackage{amssymb}
\usepackage{float}
\usepackage{subfigure}
\usepackage{threeparttable}
\usepackage{framed}
\usepackage{overpic}
\usepackage{multirow}
\usepackage{array}

\cvprfinalcopy

\providecommand{\keywords}[1]{\textbf{\textit{Keywords: }} #1}
\begin{document}

\title{MORAN: A Multi-Object Rectified Attention Network
\\for Scene Text Recognition}

\author{Canjie Luo\footnotemark[2]{ }, Lianwen Jin\thanks{Corresponding author}{ }  \footnotemark[2]{ } \footnotemark[3]{ },  Zenghui Sun\footnotemark[2]{ }\\
School of Electronic and Information Engineering, South China University of Technology\footnotemark[2]\\
SCUT-Zhuhai Institute of Modern Industrial Innovation\footnotemark[3]\\
{\tt\small \{canjie.luo, lianwen.jin\footnotemark[1] , sunfreding\}@gmail.com, eelwjin@scut.edu.cn\footnotemark[1] }}

\maketitle

\begin{abstract}
Irregular text is widely used. However, it is considerably difficult to recognize because of its various shapes and distorted patterns. In this paper, we thus propose a \textbf{m}ulti-\textbf{o}bject \textbf{r}ectified \textbf{a}ttention \textbf{n}etwork (MORAN) for general scene text recognition. The MORAN consists of a multi-object rectification network and an attention-based sequence recognition network. The multi-object rectification network is designed for rectifying images that contain irregular text. It decreases the difficulty of recognition and enables the attention-based sequence recognition network to more easily read irregular text. It is trained in a weak supervision way, thus requiring only images and corresponding text labels. The attention-based sequence recognition network focuses on target characters and sequentially outputs the predictions. Moreover, to improve the sensitivity of the attention-based sequence recognition network, a fractional pickup method is proposed for an attention-based decoder in the training phase. With the rectification mechanism, the MORAN can read both regular and irregular scene text. Extensive experiments on various benchmarks are conducted, which show that the MORAN achieves state-of-the-art performance. The source code is available\footnote{\url{https://github.com/Canjie-Luo/MORAN_v2}}.
\end{abstract}


\section{Introduction}
{S}{cene} text recognition is an essential process in computer vision tasks. Many practical applications such as traffic sign reading, product recognition, intelligent inspection, and image searching, benefit from the rich semantic information of scene text. With the development of scene text detection methods \cite{gomez2017textproposals,khare2016blind,sun2015robust,zhu2016could}, scene character recognition has emerged at the forefront of this research topic and is regarded as an open and very challenging research problem \cite{su2017accurate}.

\begin{figure}[t]
\centering
\subfigure[]{
\begin{minipage}[c]{0.15\textwidth}
\centering
  \includegraphics[width=2.5cm,height=1.5cm]{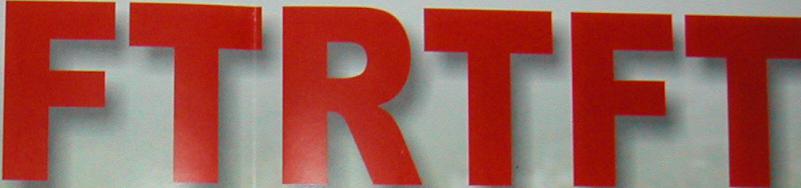}
  \includegraphics[width=2.5cm,height=1.5cm]{picture/regular-b.jpg}
  \includegraphics[width=2.5cm,height=1.5cm]{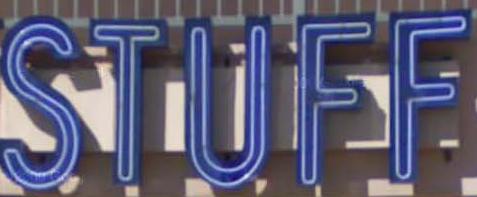}
\end{minipage}%
}%
\subfigure[]{
\begin{minipage}[c]{0.15\textwidth}
\centering
  \includegraphics[width=2.5cm,height=1.5cm]{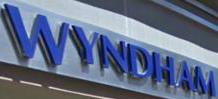}
  \includegraphics[width=2.5cm,height=1.5cm]{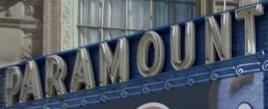}
  \includegraphics[width=2.5cm,height=1.5cm]{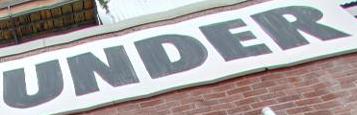}
\end{minipage}%
}%
\subfigure[]{
\begin{minipage}[c]{0.15\textwidth}
\centering
  \includegraphics[width=2.5cm,height=1.5cm]{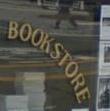}
  \includegraphics[width=2.5cm,height=1.5cm]{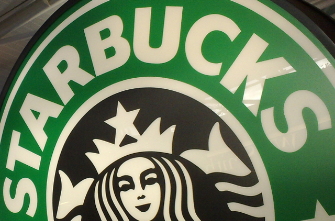}
  \includegraphics[width=2.5cm,height=1.5cm]{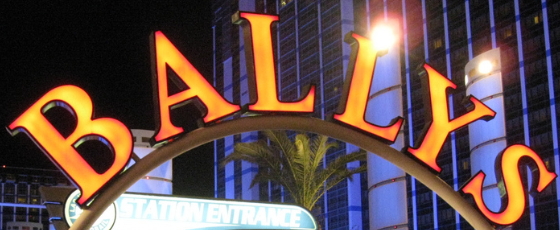}
\end{minipage}%
}%

\caption{Examples of regular and irregular scene text. (a) Regular text. (b) Slanted and perspective text. (c) Curved text.}
\label{fig:1-scene-text}
\end{figure}

Nowadays, regular text recognition methods \cite{bissacco2013photoocr,neumann2012real,shi2017end,su2017accurate,wang2012end} have achieved notable success. Moreover, methods based on convolutional neural networks \cite{bissacco2013photoocr,jaderberg2016reading,wang2012end} have been broadly applied. Integrating recognition models with recurrent neural networks \cite{he2016reading,shi2017end,shi2016robust} and attention mechanisms \cite{cheng2017focusing,cheng2017arbitrarily,lee2016recursive,yang2017learning} yields better performance for these models.

Nevertheless, most current recognition models remain too unstable to handle multiple disturbances from the environment. Furthermore, the various shapes and distorted patterns of irregular text cause additional challenges in recognition. As illustrated in Fig.  \ref{fig:1-scene-text}, scene text with irregular shapes, such as perspective and curved text, is still very challenging to recognize.

\begin{figure}
\centering
\begin{overpic}[width=8.5cm,height=3.2cm]{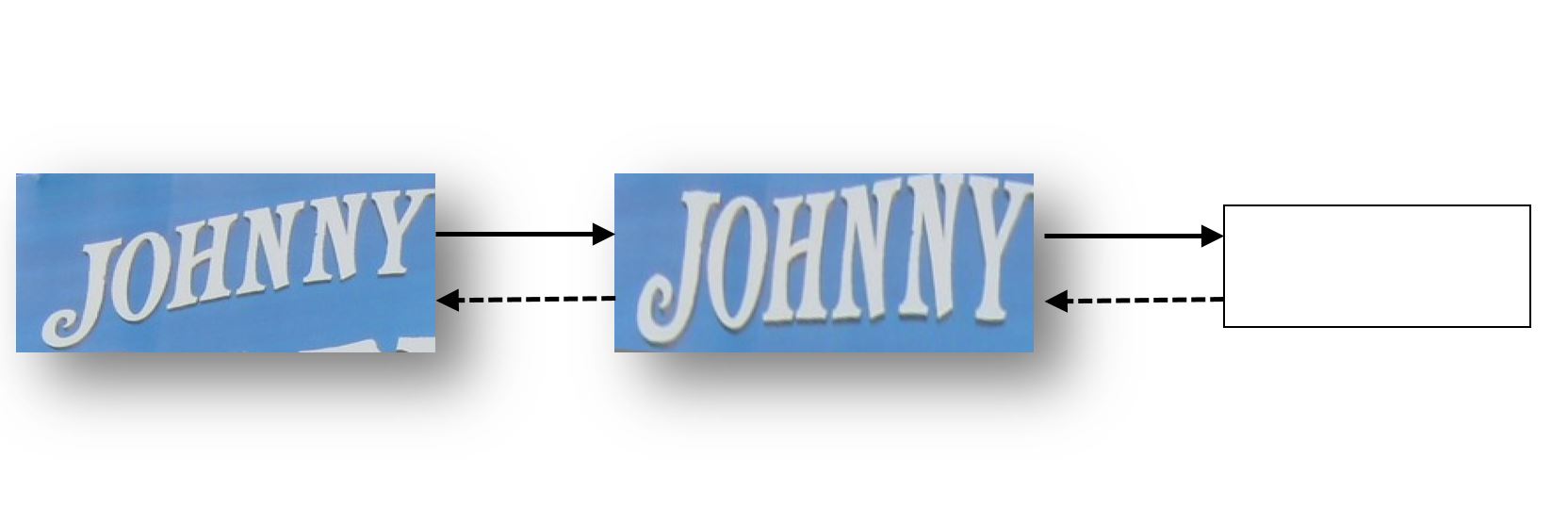}

\put(11,33){Input}
\put(10,28){Image}
\put(45,33){Rectified}
\put(48,28){Image}
\put(82,26){Result}

\put(79,17){JOHNNY}
\put(28,26){MORN}
\put(66,26){ASRN}

\put(29,7){Weak}
\put(25,3){Supervision}

\put(65,7){Text Label}
\put(64,3){Supervision}

\end{overpic}

\caption{Overview of the MORAN. The MORAN contains a MORN and an ASRN. The image is rectified by the MORN and given to the ASRN. The dashed lines show the direction of gradient propagation, indicating that the two sub-networks are jointly trained.}
\label{fig:2-MORAN-system}
\end{figure}
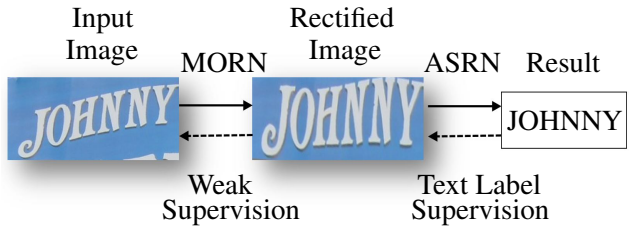

Reading text is naturally regarded as a multi-classification task involving sequence-like objects \cite{shi2017end}. Usually, the characters in one text are of the same size. However, characters in different scene texts can vary in size. Therefore, we propose the multi-object rectified attention network (MORAN), which can read rotated, scaled and stretched characters in different scene texts. The MORAN consists of a multi-object rectification network \textbf{(MORN)} to rectify images and an attention-based sequence recognition network \textbf{(ASRN)} to read the text. We separate the difficult recognition task into two parts. First, as one kind of spatial transformer, the MORN rectifies images that contain irregular text. As Fig. \ref{fig:2-MORAN-system} shows, after the rectification by the MORN, the slanted text becomes more horizontal, tightly-bounded, and easier to read. Second, ASRN takes the rectified image as input and outputs the predicted word.

The training of the MORN is guided by the ASRN, which requires only text labels. Without any geometric-level or pixel-level supervision, the MORN is trained in a weak supervision way. To facilitate this manner of network training, we initialize a basic coordinate grid. Every pixel of an image has its own position coordinates. The MORN learns and generates an offset grid based on these coordinates and samples the pixel value accordingly to rectify the image. The rectified image is then obtained for the ASRN.

With respect to the ASRN, a decoder with an attention mechanism is more likely to predict the correct words because of the rectified images. However, Cheng et al. \cite{cheng2017focusing} found that existing attention-based methods cannot obtain accurate alignments between feature areas and targets. Therefore, we propose a fractional pickup method to train the ASRN. By adopting several scales of stretch on different parts of the feature maps, the feature areas are changed randomly at every iteration in the training phase. Owing to training with fractional pickup, the ASRN is more robust to the variation of context. Experiments show that the ASRN can accurately focus on objects.

In addition, we designed a curriculum learning strategy for the training of the MORAN. Because the MORN and ASRN are mutually beneficial in terms of performance, we first fix one of them to more efficiently optimize the other. Finally, the MORN and ASRN are optimized in an end-to-end fashion to improve performance. In short, the contributions of our research are as follows:

\begin{itemize}

\item We propose the MORAN framework to recognize irregular scene text. The framework contains a multi-object rectification network (MORN) and an attention-based sequence recognition network (ASRN). The image rectified by the MORN is more readable for the ASRN.

\item Trained in a weak supervision way, the sub-network MORN is flexible. It is free of geometric constraints and can rectify images with complicated distortion.

\item We propose a fractional pickup method for the training of the attention-based decoder in the ASRN. To address noise perturbations, we expand the visual field of the MORAN, which further improves the sensitivity of the attention-based decoder.

\item We propose a curriculum learning strategy that enables the MORAN to learn efficiently. Owing to the training with this strategy, the MORAN outperforms state-of-the-art methods on several standard text recognition benchmarks, including the IIIT5K, SVT, ICDAR2003, ICDAR2013, ICDAR2015, SVT-Perspective, and CUTE80 datasets.

\end{itemize}

The rest of the paper is organized as follow. Section 2 reviews related work. Section 3 details the proposed method. Experimental results are given in Section 4, and the conclusions are presented in Section 5.

\section{Related Work}
\label{section:Related work}
In recent years, the recognition of scene text has greatly advanced because of the rapid development of neural networks \cite{gu2017recent}. Zhu et al. \cite{zhu2016scene} and Ye et al. \cite{ye2015text} have provided an overview of the major advances in the field of scene text detection and recognition. Based on the sliding window method \cite{wang2011end,wang2010word}, pattern features extracted by a neural network become dominant with respect to the hand crafted features, such as the connected components \cite{neumann2012real}, strokelet generation \cite{yao2014strokelets}, histogram of oriented gradients descriptors \cite{dalal2005histograms,su2014accurate}, tree-structured models \cite{shi2014end}, semi-markov conditional random fields \cite{seok2015scene} and generative shape models \cite{lou2016generative}. For instance, Bissacco \cite{bissacco2013photoocr} applied a network with five hidden layers for character classification. Using convolutional neural networks (CNNs), Jaderberg et al. \cite{Jaderberg2015Deep} and Yin et al. \cite{yin2017scene} proposed respective methods for unconstrained recognition.

With the widespread application of recurrent neural networks (RNNs)  \cite{cho2014learning,hochreiter1997long}, CNN-based methods are combined with RNNs for better learning of context information. As a feature extractor, the CNN obtains the spatial features of images. Then, the RNN learns the context of features. Shi et al. \cite{shi2017end} proposed an end-to-end trainable network with both CNNs and RNNs, named CRNN. Guided by the CTC loss \cite{graves2006connectionist}, the CRNN-based network learns the conditional probability between predictions and sequential labels.

Furthermore, attention mechanisms \cite{bahdanau2014neural} focus on informative regions to achieve better performance. Lee et al. \cite{lee2016recursive} proposed a recursive recurrent network with attention modeling for scene text recognition. Yang et al. \cite{yang2017learning} addressed a two-dimensional attention mechanism. Cheng et al. \cite{cheng2017focusing} used the focusing attention network (FAN) to correct shifts in attentional mechanisms and achieved more accurate position predictions.

Compared with regular text recognition work, irregular text recognition is more difficult. One kind of irregular text recognition method is the bottom-up approach \cite{cheng2017arbitrarily,yang2017learning}, which searches for the position of each character and then connects them. Another is the top-down approach \cite{liu2016star,shi2016robust}. This type of approach matches the shape of the text, attempts to rectify it, and reduces the degree of recognition difficulty.

In the bottom-up manner, a two-dimensional attention mechanism for irregular text was proposed by Yang et al. \cite{yang2017learning}. Based on the sliced Wasserstein distance \cite{rabin2011wasserstein}, the attention alignment loss is adopted in the training phase, which enables the attention model to accurately extract the character features while ignoring the redundant background information. Cheng et al. \cite{cheng2017arbitrarily} proposed an arbitrary-orientation text recognition network, which uses more direct information of the position to instruct the network to identify characters in special locations.

In the top-down manner, STAR-Net \cite{liu2016star} used an affine transformation network that transforms the rotated and differently scaled text into more regular text. Meanwhile, a ResNet \cite{he2016deep} is used to extract features and handle more complex background noise. RARE \cite{shi2016robust} regresses the fiducial transformation points on sloped text and even curved text, thereby mapping the corresponding points onto standard positions of the new image. Using thin-plate-spline \cite{bookstein1989principal} to back propagate the gradients, RARE is end-to-end optimized.

Our proposed MORAN model uses the top-down approach. The fractional pickup training method is thus designed to improve the sensitivity of the MORAN to focus on characters. For the training of the MORAN, we propose a curriculum learning strategy for better convergence.

\section{Methodology}

The MORAN contains two parts. One is the MORN, which is trained in a weak supervision way to learn the offset of each part of the image. According to the predicted offsets, we apply sampling and obtain a rectified text image. The other one is ASRN, a CNN-LSTM framework followed by an attention decoder. The proposed fractional pickup further improves attention sensitivity. The curriculum learning strategy guides the MORAN to achieve state-of-the-art performance.

\subsection{Multi-Object Rectification Network}
\label{section:Multi-Object Rectification Network}
\begin{figure}
\centering
\includegraphics[width=5cm,height=5cm]{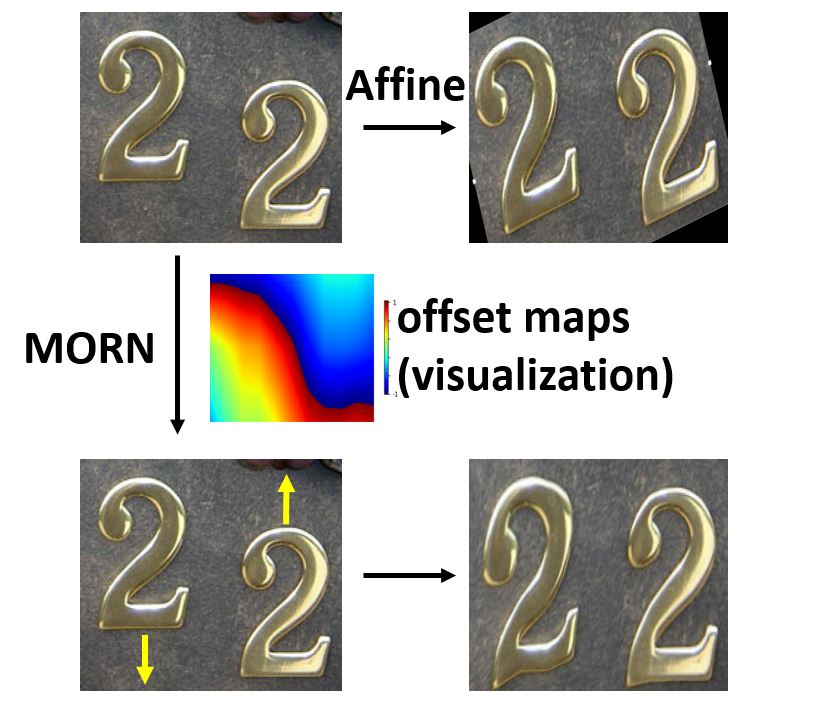}
\caption{Comparison of the MORN and affine transformation. The MORN is free of geometric constraints. The main direction of rectification predicted by the MORN for each character is indicated by a yellow arrow. The offset maps generated by the MORN are visualized as a heat map. The offset values on the boundary between red and blue are zero. The directions of rectification on both sides of the boundary are opposite and outward. The depth of the color represents the magnitude of the offset value. The gradual-change in color indicates the smoothness of the rectification.}
\label{fig:compare-stn}
\end{figure}

Common methods to rectify patterns such as the affine transformation network, are limited by certain geometric constraints. With respect to the affine transformation, it is limited to rotation, scaling, and translation. However, one image may have several kinds of deformations, and the distortion of scene text will thus be complicated. As shown in Fig. \ref{fig:compare-stn}, the characters in the image become slanted after rectification by the affine transformation. The black edges introduce additional noise. Therefore, transformations with geometric constraints can not cover all complicated deformations.

Another method that is free of geometric constraints, is the deformable convolutional network \cite{deformable2017}. Using deformable convolutional kernels, the feature extractor automatically selects informative features. We attempted to combine the recognition network with a deformable convolutional network. However, as a sequence-to-sequence problem, irregular text recognition is more challenging. The network sometimes failed to converge. The best accuracy rate on IIIT5K we achieved was only 78.1\%, which is far behind the state-of-the-art result (91.2\%).

\begin{figure*}[t]
\centering
\includegraphics[width=17cm]{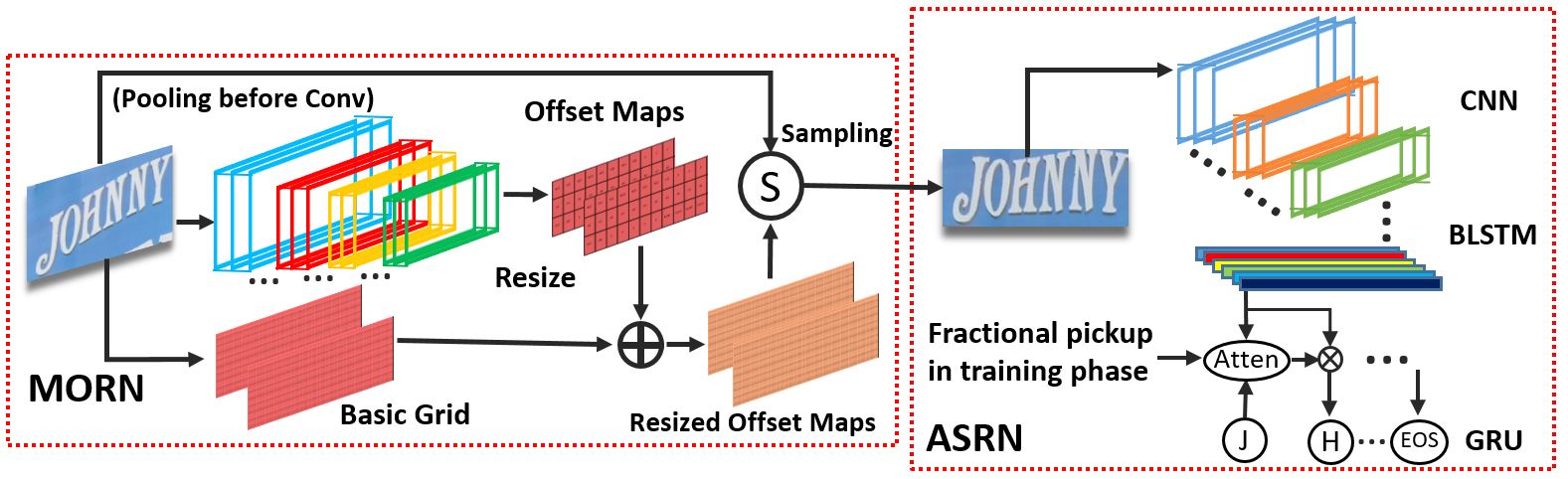}
\caption{Overall structure of MORAN. }
\label{fig:3-MORAN-overview}
\end{figure*}

Because the recognition models remain inadequately strong to handle multiple disturbances from various shapes, we consider rectifying images to reduce the difficulty of the recognition. As demonstrated in Fig. \ref{fig:3-MORAN-overview}, the MORN architecture rectifies the distorted image. The MORN predicts the offset of each part of the image without any geometric constraint. Based on the predicted offsets, the image is rectified and becomes easier to recognize.

Furthermore, the MORN predicts the position offsets but not the categories of characters. The character details for classification are not necessary. We hence place a pooling layer before the convolutional layer to avoid noise and reduce the amount of calculation.

\begin{table}[h]
\centering
\caption{Architecture of the MORN}
\label{table:The architecture of MORN}
\begin{tabular*}{8.5cm}{c|p{3.5cm}<{\centering}|c}
\hline
Type & Configurations  & Size \\
\hline
Input & - & 1$\times$32$\times$100 \\
\hline
MaxPooling & k2, s2 & 1$\times$16$\times$50 \\
\hline
Convolution  & maps:64, k3, s1, p1 & 64$\times$16$\times$50 \\
\hline
MaxPooling & k2, s2 & 64$\times$8$\times$25 \\
\hline
Convolution  & maps:128, k3, s1, p1 & 128$\times$8$\times$25 \\
\hline
MaxPooling & k2, s2 & 128$\times$4$\times$12 \\
\hline
Convolution  & maps:64, k3, s1, p1 & 64$\times$4$\times$12 \\
\hline
Convolution  & maps:16, k3, s1, p1 & 16$\times$4$\times$12 \\
\hline
Convolution  & maps:2, k3, s1, p1 & 2$\times$4$\times$12 \\
\hline
MaxPooling & k2, s1 & 2$\times$3$\times$11 \\
\hline
Tanh & - & 2$\times$3$\times$11 \\
\hline
Resize & - & 2$\times$32$\times$100 \\
\hline
\end{tabular*}
\begin{tablenotes}
\item Here k, s, p are kernel, stride and
padding sizes, respectively. For example, $k3$ represents a $3\times3$ kernel size.
\end{tablenotes}
\end{table}

The architecture of the MORN is given in Table\ref{table:The architecture of MORN}. Each convolutional layer is followed by a batch normalization layer and a ReLU layer except for the last one. The MORN first divides the image into several parts and then predicts the offset of each part. With an input size of $32\times100$, the MORN divides the image into $3\times 11 = 33$ parts. All the offset values are activated by $Tanh(\cdot)$, resulting in values within the range of $(-1, 1)$. The offset maps contain two channels, which denote the x-coordinate and y-coordinate respectively. Then, we apply bilinear interpolation to smoothly resize the offset maps to a size of $32\times100$, which is the same size of the input image. After allocating the specific offset to each pixel, the transformation of the image is smooth. As demonstrated in Fig.\ref{fig:compare-stn}, the color depth gradually changes on both sides of the boundary between the red and blue colors in the heat map, which evidences the smoothness of the rectification. There are no indented edges in the rectified image.

Moreover, because every value in the offset maps represents the offset from the original position, we generate a basic grid from the input image to represent the original positions of the pixels. The basic grid is generated by normalizing the coordinate of each pixel to $[-1, 1]$. The coordinates of the top-left pixel are $(-1, -1)$, and those of the bottom-right one are $(1, 1)$. Pixels at the same positions on different channels have the same coordinates. Similar to the offset maps, the grid contains two channels, which represent the x-coordinate and y-coordinate, respectively. Then, the basic grid and the resized offset maps are summed as follows,
\begin{equation}
offset_{(c,i,j)}^{'} = offset_{(c,i,j)}+basic_{(c,i,j)} ,  c = 1,2
\end{equation}
where $(i,j)$ is the position of the $i$-th row and $j$-th column.

Before sampling, the x-coordinate and y-coordinate on the offset maps are normalized to $[0, W]$ and $[0, H]$, respectively. Here, $H\times W$ is the size of the input image. The pixel value of $i$-th row and $j$-th column in rectified image $I'$ is,
\begin{equation}
I'_{(i, j)} = I_{(i^{'}, j^{'})} \label{sampling}
\end{equation}
\begin{equation}
\left\{
\begin{aligned}
i^{'} = offset_{(1, i, j)}^{'} \\
j^{'} = offset_{(2, i, j)}^{'}
\end{aligned}
\right.
\end{equation}
where $I$ is the input image. Further, $i^{'}$ is obtained from the first channel of the offset maps, whereas $j^{'}$ is from the second channel. Both $i^{'}$ and $j^{'}$ are real values as opposed to integers so rectified image $I'$ is sampled from $I$ using bilinear interpolation.

Because Equation (\ref{sampling}) is differentiable, the MORN can back-propagate the gradients. The MORN can be trained in a weak supervision way with images and associated text labels only, which means that it does not need pixel-level labeling information about the deformation of the text.

\begin{figure}[h]
\centering
\subfigure[Perspective texts]{
\begin{minipage}[c]{0.45\textwidth}
\centering
  \includegraphics[width=7cm,height=4cm]{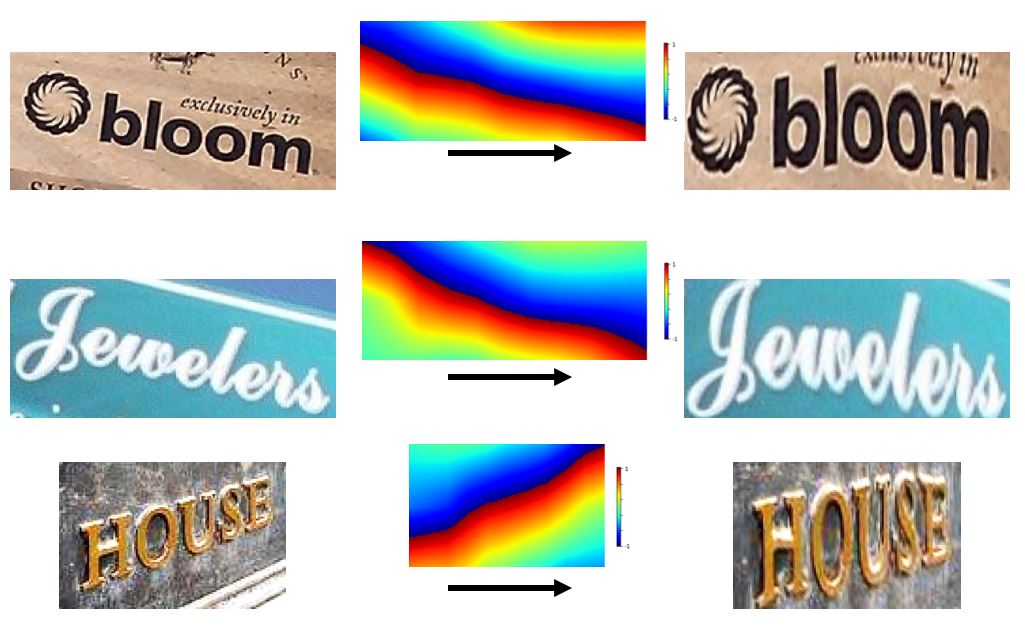}
\end{minipage}%
}%

\subfigure[curved texts]{
\begin{minipage}[c]{0.45\textwidth}
\centering
  \includegraphics[width=7cm,height=4cm]{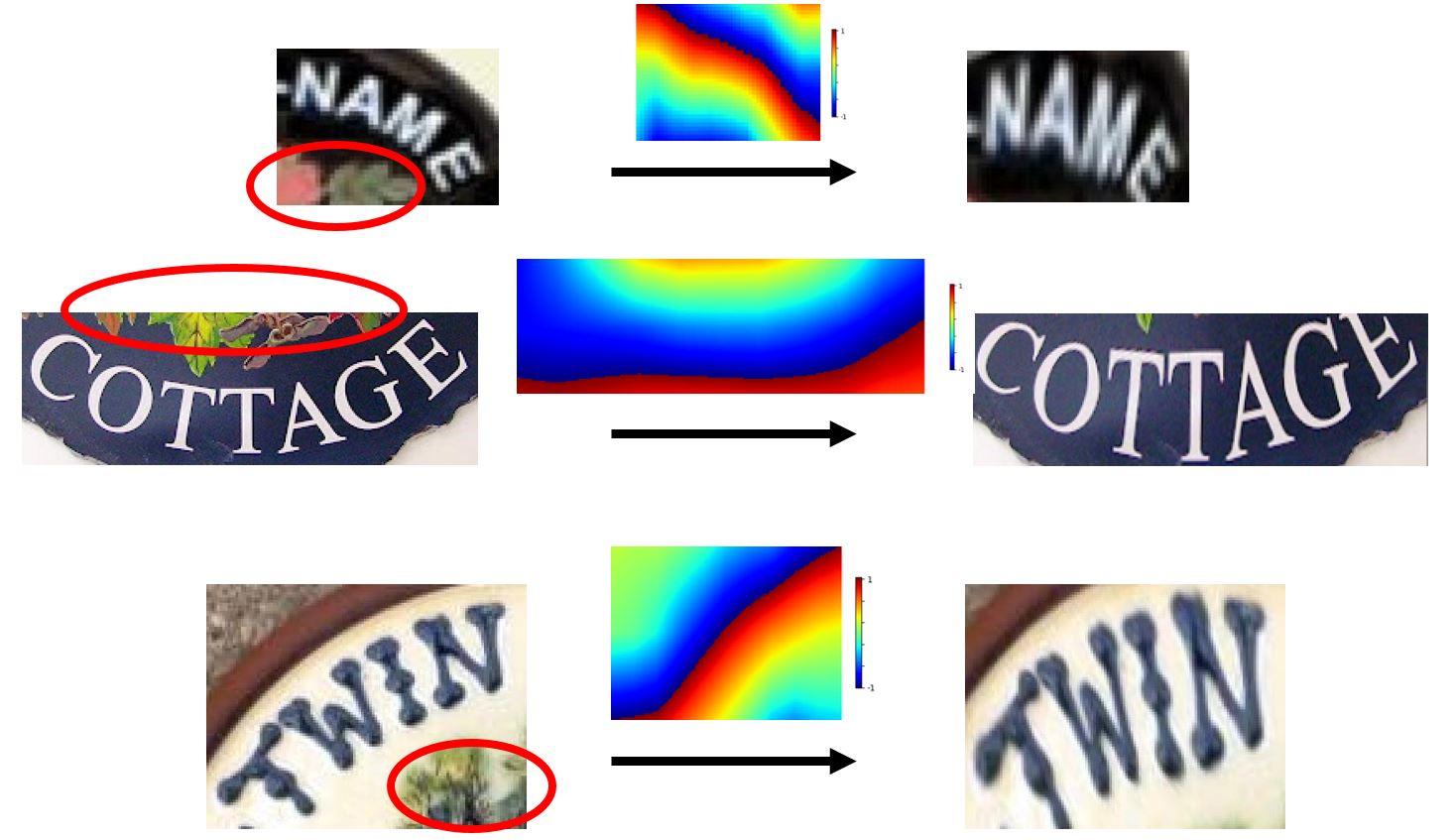}
\end{minipage}%
}%

\caption{Results of the MORN on challenging image text. The input images are shown on the left and the rectified images are shown on the right. The heat maps visualize offset maps as well as Fig. \ref{fig:compare-stn}. (a) Slanted and perspective text. (b) Curved text, which is more challenging for recognition. Removed background textures are indicated by red circles.}
\label{fig:4-ori-rectified-img}
\end{figure}

As Fig. \ref{fig:4-ori-rectified-img} shows, the text in the input images is irregular. However, the text in the rectified images is more readable. Slanted or perspective texts become tightly bound after rectification. Furthermore, redundant noise is eliminated by the MORN for the curved texts. The background textures are removed in the rectified images of Fig. \ref{fig:4-ori-rectified-img} (b).

The advantages of the MORN are manifold. 1) The rectified images are more readable owing to the regular shape of the text and the reduced noise. 2) The MORN is more flexible than the affine transformation. It is free of geometric constraints, which enables it to rectify images using complicated transformations. 3) The MORN is more flexible than methods using a specific number of regressing points. Existing method \cite{shi2016robust} cannot capture the text shape in details if the width of the image is large. Thus the MORN has no limit with respect to the image size, especially the width of the input image. 4) The MORN does not require extra labelling information of character positions. Therefore, it can be trained in a weak supervision way by using existing training datasets.

\subsection{Attention-based Sequence Recognition Network}
\label{section:Attention-based Sequence Recognition Network}
As Fig. \ref{fig:3-MORAN-overview} shows, the major structure of the ASRN is a CNN-BLSTM framework. We adopt a one-dimensional attention mechanism at the top of CRNN. The attention-based decoder, proposed by Bahdanau et al. \cite{bahdanau2014neural}, is used to accurately align the target and label. It is based on an RNN and directly generates the target sequence $(y_{1},y_{2} ...,y_{N})$. The largest number of steps that the decoder generates is $T$. The decoder stops processing when it predicts an end-of-sequence token $``EOS"$ \cite{sutskever2014sequence}. At time step $t$, output $y_t$ is,
\begin{equation}
y_{t} = Softmax(W_{out}s_{t}+b_{out})
\end{equation}
where $s_{t}$ is the hidden state at time step $t$. We update $s_{t}$ using GRU \cite{cho2014learning}. State $s_{t}$ is computed as:
\begin{equation}
s_t = GRU(y_{prev}, g_{t}, s_{t-1})
\end{equation}
where $y_{prev}$ denotes the embedding vectors of the previous output $y_{t-1}$ and $g_{t}$ represents the glimpse vectors, respectively calculated as,
\begin{equation}
y_{prev} = Embedding(y_{t-1})
\end{equation}
\begin{equation}
g_{t} = \sum_{i=1}^L(\alpha_{t,i} h_{i}) \label{equ-g}
\end{equation}
where $h_{i}$ denotes the sequential feature vectors and $L$ is the length of the feature maps. In addition, $\alpha_{t,i}$ is the vector of attention weights as follows,
\begin{equation}
\alpha_{t,i} = {exp(e_{t,i})}  /  {\sum_{j=1}^L(exp(e_{t,j}))} \label{equ-alpha}
\end{equation}
\begin{equation}
e_{t,i} = Tanh(W_{s}s_{t-1}+W_{h}h_{i}+b) \label{equ-e}
\end{equation}

Here, $W_{out}$, $b_{out}$, $W_{s}$, $W_{h}$ and $b$ are trainable parameters. Note that $y_{prev}$ is embedded from the ground truth of the last step in the training phase, whereas the ASRN only uses the predicted output of the last step as $y_{t-1}$ in the testing phase.

The decoder outputs the predicted word in an unconstrained manner in lexicon-free mode. If lexicons are available, we evaluate the probability distributions for all words and choose the word with the highest probability as the final result.

The architecture of the ASRN is given in Table\ref{table:The architecture of ASRN}. Each convolutional layer is followed by a batch normalization layer and a ReLU layer.
\begin{table}[t]
\centering
\caption{Architecture of the ASRN}
\label{table:The architecture of ASRN}
\begin{tabular*}{8.5cm}{c|p{3.5cm}<{\centering}|c}
\hline
Type & Configurations  & Size \\
\hline
Input & - & 1$\times$32$\times$100 \\
\hline
Convolution  & maps:64, k3, s1, p1 & 64$\times$32$\times$100 \\
\hline
MaxPooling & k2, s2 & 64$\times$16$\times$50 \\
\hline
Convolution  & maps:128, k3, s1, p1 & 128$\times$16$\times$50 \\
\hline
MaxPooling & k2, s2 & 128$\times$8$\times$25 \\
\hline
Convolution  & maps:256, k3, s1, p1 & 256$\times$8$\times$25 \\
\hline
Convolution  & maps:256, k3, s1, p1 & 256$\times$8$\times$25 \\
\hline
MaxPooling & k2, s2x1, p0x1 & 256$\times$4$\times$26 \\
\hline
Convolution  & maps:512, k3, s1, p1 & 512$\times$4$\times$26 \\
\hline
Convolution  & maps:512, k3, s1, p1 & 512$\times$4$\times$26 \\
\hline
MaxPooling & k2, s2x1, p0x1 & 512$\times$2$\times$27 \\
\hline
Convolution  & maps:512, k2, s1 & 512$\times$1$\times$26 \\
\hline
BLSTM & hidden unit:256 & 256$\times$1$\times$26 \\
\hline
BLSTM & hidden unit:256 & 256$\times$1$\times$26 \\
\hline
GRU & hidden unit:256 & 256$\times$1$\times$26 \\
\hline
\end{tabular*}
\begin{tablenotes}
\item Here, k, s, p are kernel, stride and
padding sizes, respectively. For example, $s2\times1$ represents a $2\times1$ stride size. ``BLSTM" stands for bidirectional-LSTM. ``GRU" is in attention-based decoder.
\end{tablenotes}
\end{table}

\subsection{Fractional Pickup}

The decoder in the ASRN learns the matching relationship between labels and target characters in images. It is a data-driven process. The ability to choose regions that are focus-worthy is enhanced by the feedback of correct alignment.

However, scene text is surrounded by various types of noise. Often, the decoder is likely to be deceived into focusing on ambiguous background regions in practical applications. If the decoder generates an incorrect region of focus, the non-corresponding features are chosen, which can cause a failed prediction.

\begin{figure}[t]
\centering
\rule{8cm}{0.05em}
\\
-------------------------------------------------------------
\begin{overpic}[width=8cm,height=3cm]{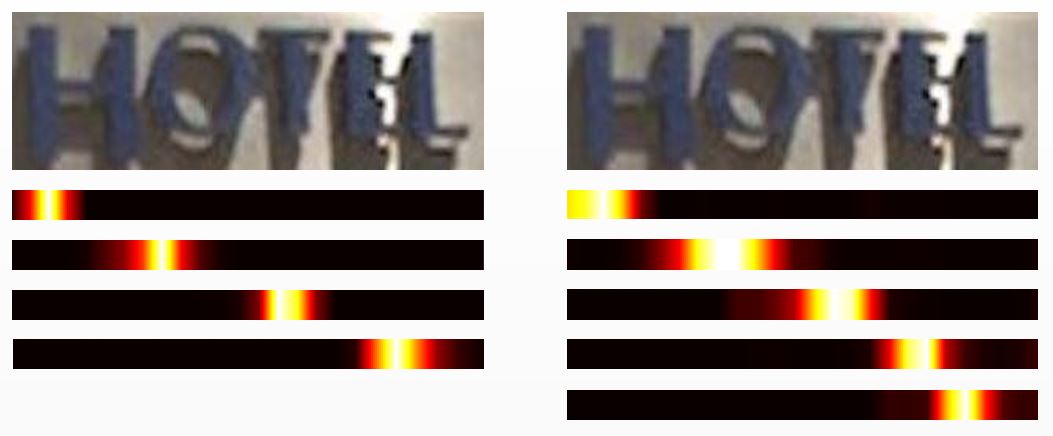}

\put(13,39){Without FP}
\put(67,39){With FP}

\put(20,-3){\color{blue}{hot}\color{red}{l}}
\put(72,-3){\color{blue}{hotel}}
\end{overpic}

-------------------------------------------------------------

\begin{overpic}[width=8cm,height=3cm]{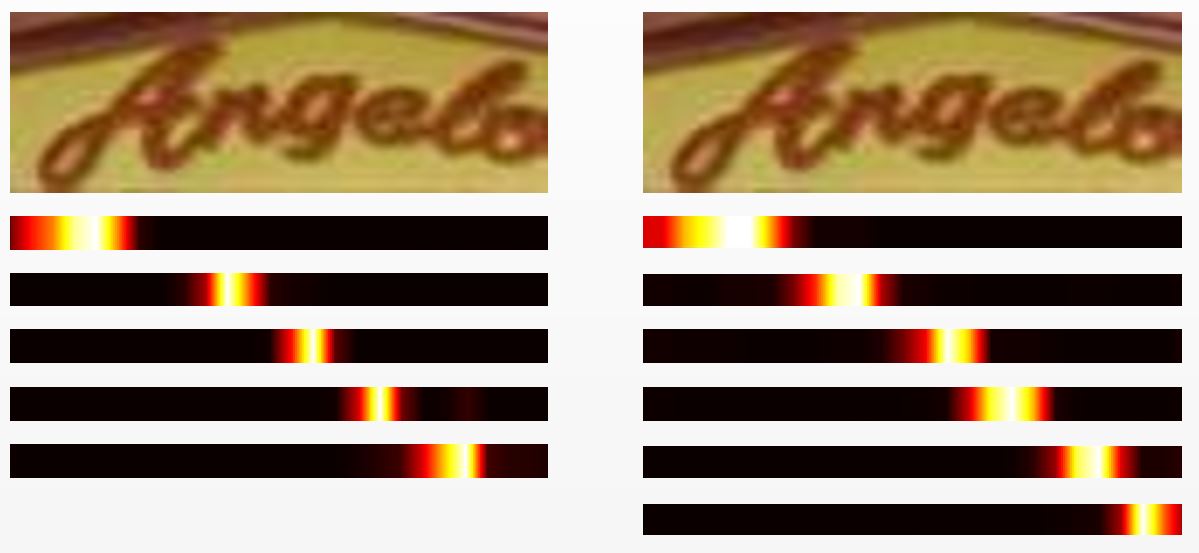}
\put(20,-3){\color{blue}{a}\color{red}{r}\color{blue}{ge}\color{red}{h}}
\put(72,-3){\color{blue}{angels}}
\end{overpic}

-------------------------------------------------------------

\begin{overpic}[width=8cm,height=3cm]{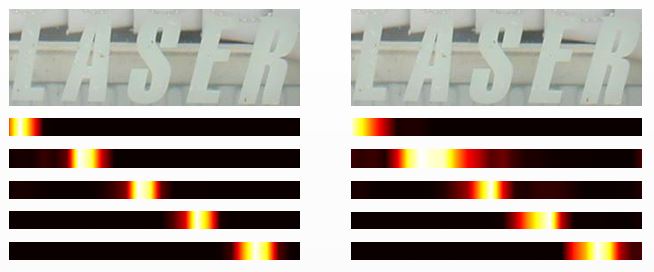}
\put(20,-3){\color{red}{e}\color{blue}{aser}}
\put(72,-3){\color{blue}{laser}}
\end{overpic}

-------------------------------------------------------------

\caption{Difference in $\alpha_{t}$ for training with and without fractional pickup. Here $\alpha_{t}$ is visualized as a heat map. We delete the $\alpha_{t}$ corresponding to ``EOS".}
\label{fig:5-fractional-pickup}
\end{figure}
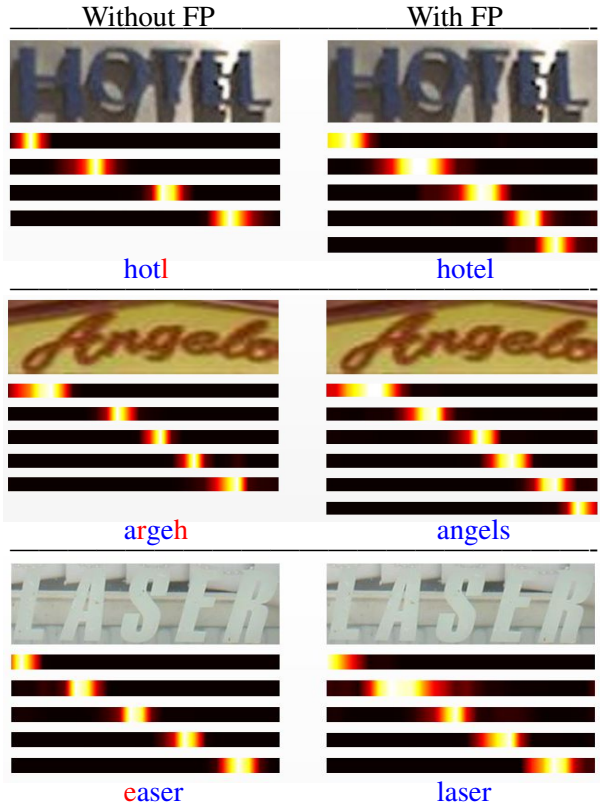

Some challenging samples for recognition are presented in Fig. \ref{fig:5-fractional-pickup}. In this figure, the images contain text with shadows and unclear boundaries between characters or complicated backgrounds. Moreover, the focus regions generated by the decoder are narrow, which increases the probability of drifting from the correct regions.

We propose a training method called fractional pickup that fractionally picks up the neighboring features in the training phase. An attention-based decoder trained by fractional pickup method can perceive adjacent characters. The wider field of attention contributes to the robustness of the MORAN.

We hence adopt fractional pickup at each time step of the decoder. In other words, a pair of attention weights are selected and modified at every time step. At time step $t$, $\alpha_{t,k}$ and $\alpha_{t,k+1}$ are updated as,
\begin{equation}
\left\{
\begin{aligned}
\alpha_{t,k}^{,} = \beta \alpha_{t,k}+(1-\beta)\alpha_{t,k+1} \\
\alpha_{t,k+1}^{,} = (1-\beta) \alpha_{t,k}+\beta\alpha_{t,k+1}
\end{aligned}
\right.
\end{equation}
where decimal $\beta$ and integer $k$ are randomly generated as,
\begin{equation}
\beta = rand(0,1)
\end{equation}
\begin{equation}
k = rand[1,T-1]
\end{equation}
Here, T is the maximum number of steps of the decoder.

\textbf{Variation of Distribution}
Fractional pickup adds randomness to $\alpha_{t,k}$ and $\alpha_{t,k+1}$ in the decoder. This means that, even for the same image, the distribution of $\alpha_{t}$ changes every time step in the training phase. As noted in Equation (\ref{equ-g}), the glimpse vectors $g_{t}$ grabs the sequential feature vectors $h_{i}$ according to the various distributions of $\alpha_{t}$, which is equivalent to the changes in feature areas. The randomness of $\beta$ and $k$ avoids over-fitting and contributes to the robustness of the decoder.

\begin{figure}
\centering
\begin{overpic}[width=4cm]{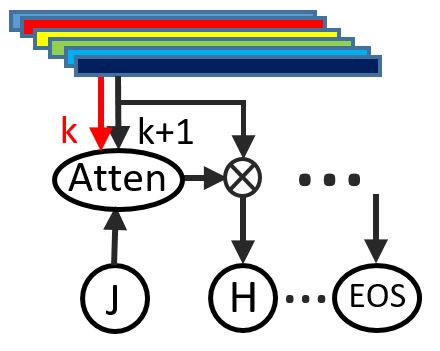}
\put(-5,65){$h_{i}$}
\end{overpic}
\caption{Fractional pickup creates a shortcut of forward propagation. The shortcut is drawn as a red arrow.}
\label{fig:short-cut}
\end{figure}

\textbf{Shortcut of Forward Propagation}
Sequential feature vector $h_{i}$ is the output of the last bidirectional-LSTM in the ASRN. As shown in Fig. \ref{fig:short-cut}, for step $k+1$ in the bidirectional-LSTM, a shortcut connecting to step $k$ is created by fractional pickup. The shortcut retains some features of the previous step in the training phase, which is the interference to the forget gate in bidirectional-LSTM. Therefore, fractional pickup provides more information about the previous step and increases the robustness for the bidirectional-LSTM in the ASRN.

\textbf{Broader Visual Field}
Training with fractional pickup disturbs the decoder through the local variation of $\alpha_{t,k}$ and $\alpha_{t,k+1}$. Note that $\alpha_{t,k}$ and $\alpha_{t,k+1}$ are neighbors. Without fractional pickup, the error term of sequence feature vector $h_{k}$ is,
\begin{equation}
\delta_{h_{k}} = \delta_{g_{t}}\alpha_{t,k}
\end{equation}
where $\delta_{g_{t}}$ is the error term of glimpse vector $g_{t}$. $\delta_{h_{k}}$ is only relevant to $\alpha_{t,k}$. However, with fractional pickup, the error item becomes,
\begin{equation}
\delta_{h_{k}} = \delta_{g_{t}}(\beta \alpha_{t,k}+(1-\beta)\alpha_{t,k+1})
\end{equation}
where $\alpha_{t,k+1}$ is relevant to $h_{k+1}$, as noted in Equations (\ref{equ-alpha}) and (\ref{equ-e}), which means $\delta_{h_{k}}$ is influenced by the neighbouring features. Owing to the disturbance, back-propagated gradients are able to dynamically optimize the decoder over a broader range of neighbouring regions.

The MORAN trained with fractional pickup method generates a smoother $\alpha_{t}$ at each time step. Accordingly, it extracts features not only of the target characters, but also of the foreground and background context. As demonstrated in Fig. \ref{fig:5-fractional-pickup}, the expanded visual field enables the MORAN to correctly predict target characters. To the best of our knowledge, this is the first attempt to adopt a shortcut in the training of the attention mechanism.

\subsection{Curriculum Training}
\label{section:curriculum-training}

The MORAN is end-to-end trainable with random initialization. However, end-to-end training consumes considerable time. We found that the MORN and ASRN can hinder each other during training. A MORN cannot be guided to rectify images when the input images have been correctly recognized by the high-performance ASRN. For the same reason, the ASRN will not gain robustness because the training samples have already been rectified by the MORN. The reasons above lead to inefficient training.

Therefore, we propose a curriculum learning strategy to guide each sub-network in MORAN. The strategy is a three-step process. We first optimize the MORN and ASRN respectively and then join them together for further end-to-end training. The difficulty of training samples is gradually increased. The training set is denoted as $D = \left \{I_{i}, Y_{i} \right \}, i=1...N $. We minimize the negative log-likelihood of conditional probability of $D$ as follows:

\begin{equation}
Loss = -\sum_{i=1}^N{ \sum_{t=1}^{\left| Y_{i} \right|}{\log p(Y_{i,t} \left| \right. I_{i}; \theta)} }
\end{equation}
where $Y_{i,t}$ is the ground truth of the $t$-th character in $I_{i}$. $\theta$ denotes the parameters of MORAN.

\textbf{First Stage for ASRN}
We first optimize the ASRN by using regular training samples. The dataset released by Gupta et al. \cite{gupta2016synthetic} has tightly bounded annotations, which makes it possible to crop a text region with a tightly bounded box. The ASRN is first trained with these regular samples. Then, we simply crop every text using a minimum circumscribed horizontal rectangle to obtain irregular training samples. The commonly used datasets released by Jaderberg et al. \cite{jaderberg2014synthetic} and Gupta et al. \cite{gupta2016synthetic} offer abundant irregular training samples. We use them for the following training. Taking advantage of them, we optimize ASRN, which thus achieves higher accuracy.

\textbf{Second Stage for MORN}
The ASRN trained using regular training samples is chosen to guide the MORN training. This ASRN is not adequately robust for irregular text recognition so it is able to provide informative gradients for the MORN. We fix the parameters of this ASRN, and stack it after the MORN. If the transformation of the MORN does not reduce the difficulty of recognition, few meaningful gradients will be provided by the ASRN. The optimization of MORN would not progress. Only the correct transformation that decreases difficulty for recognition will give positive feedback to the MORN.

\textbf{Third Stage for End-to-end Optimization}
After the MORN and ASRN are optimized individually, we connect them for joint training in an end-to-end fashion. Joint training enables MORAN to complete end-to-end optimization and outperform state-of-the-art methods.

\section{Experiments}
In this section we describe extensive experiments conducted on various benchmarks, including regular and irregular datasets. The performances of all the methods are measured by word accuracy.

\begin{table*}[t]
\centering
\caption{Comparison of pooling layers in lexicon-free mode. ``No", ``AP" and ``MP" respectively indicate no pooling layer, an average-pooling layer and a max-pooling layer at the top of the MORN. The kernel size is 2. ``s" represents the stride. }
\label{table:comparison-of-pooling}
\begin{tabular}{|c| c | c | c | c | c | c | c | c | c}
\hline
\multirow{2}{*}{} & s & IIIT5K & SVT & IC03 & IC13 & SVT-P & CUTE80 & IC15 \\
\cline{2-9}
\hline
No & - & 85.7 & 87.9 & 92.9 & 91.5 & 75.8 & 65.9 & 59.4  \\
AP & 2 & 89.2 & 87.4 & 94.8 & 91.1 & 75.9 & 71.1 & 64.6 \\
AP & 1 & 89.3 & 87.9 & 94.7 & 91.6 & 75.9 & 72.9 & 64.9 \\
MP & 2 & 90.4 & 88.2 & 94.5 & 91.8 & \textbf{76.1} & 76.4 & 68.4 \\
MP & 1 & \textbf{91.2} & \textbf{88.3} & \textbf{95.0} & \textbf{92.4} & \textbf{76.1} & \textbf{77.4} & \textbf{68.8} \\
\hline
\end{tabular}
\end{table*}

\begin{table*}[t]
\centering
\caption{Performance of the MORAN. }
\label{table:Performance-of-MORAN}
\begin{tabular}{|c|c|c|c|c|c|c|c|}
\hline
\multirow{1}{*}{Method} & IIIT5K & SVT & IC03 & IC13 & SVT-P & CUTE80 & IC15 \\
\cline{2-8}
\hline
End-to-end training & 89.9 & 84.1 & 92.5 & 90.0 & 76.1 & 77.1 & 68.8 \\
\hline
Only ASRN & 84.2 & 82.2 & 91.0 & 90.1 & 71.0 & 64.6 & 65.6 \\
MORAN without FP & 89.7 & 87.3 & 94.5 & 91.5 & 75.5 & 77.1 & 68.6 \\
MORAN with FP & \textbf{91.2} & \textbf{88.3} & \textbf{95.0} & \textbf{92.4} & \textbf{76.1} & \textbf{77.4} & \textbf{68.8} \\
\hline
\end{tabular}
\end{table*}

\subsection{Datasets}

\textbf{IIIT5K-Words (IIIT5K)} \cite{mishra2012scene} contains 3000 cropped word images for testing. Every image has a 50-word lexicon and a 1000-word lexicon. The lexicon consists of a ground truth and some randomly picked words.

\textbf{Street View Text (SVT)} \cite{wang2011end} was collected from the Google Street View, consisting of 647 word images. Many images are severely corrupted by noise and blur, or have very low resolutions. Each image is associated with a 50-word lexicon.

\textbf{ICDAR 2003 (IC03)} \cite{lucas2003icdar} contains 251 scene images that are labeled with text bounding boxes. For fair comparison, we discarded images that contain non-alphanumeric characters or those have less than three characters, following Wang, Babenko, and Belongie \cite{wang2011end}. The filtered dataset contains 867 cropped images. Lexicons comprise of a 50-word lexicon defined by Wang et al. \cite{wang2011end} and a ``full lexicon". The latter lexicon combines all lexicon words.

\textbf{ICDAR 2013 (IC13)} \cite{karatzas2013icdar} inherits most of its samples from IC03. It contains 1015 cropped text images. No lexicon is associated with this dataset.

\textbf{SVT-Perspective (SVT-P)} \cite{quy2013recognizing} contains 645 cropped images for testing. Images are selected from side-view angle snapshots in Google Street View. Therefore, most images are perspective distorted. Each image is associated with a 50-word lexicon and a full lexicon.

\textbf{CUTE80} \cite{risnumawan2014robust} contains 80 high-resolution images taken in natural scenes. It was specifically collected for evaluating the performance of curved text recognition. It contains 288 cropped natural images for testing. No lexicon is associated with this dataset.

\textbf{ICDAR 2015 (IC15)} \cite{karatzas2015icdar} contains 2077 cropped images including more than 200 irregular text. No lexicon is associated with this dataset.

\subsection{Implementation Details}
\textbf{Network: }Details about the MORN and the ASRN of MORAN are given in Table\ref{table:The architecture of MORN} and Table\ref{table:The architecture of ASRN} respectively. The number of hidden units of GRU in the decoder is $256$. The ASRN outputs 37 classes, including 26 letters, 10 digits and a symbol standing for $``EOS"$.

\textbf{Training Model: }As stated in Section  \ref{section:curriculum-training}, the training of the MORAN is guided by a curriculum learning strategy. The training data consists of 8-million synthetic images released by Jaderberg et al. \cite{jaderberg2014synthetic} and 6-million synthetic images released by Gupta et al. \cite{gupta2016synthetic}. No extra data is used. We do not use any geometric-level or pixel-level labels in our experiments. Without any fine-tuning for each specific dataset, the model is trained using only synthetic text. With ADADELTA \cite{zeiler2012adadelta} optimization method, we set learning rate to $1.0$ at the beginning and decreased it to $0.01$ in the third stage of the curriculum learning strategy. Following the similar settings in \cite{liu2016star}, we found that a decreased learning rate contributes to better convergence. The batch size was set to 64. We trained the model for 600,000, 20,000 and 300,000 iterations respectively in three stages of the curriculum learning strategy. The training totally consumed 30 hours.

\textbf{Implementation: }We implemented our method under the framework of PyTorch \cite{pytorch}. CUDA 8.0 and CuDNN v7 backends are used in our experiments so our model is GPU-accelerated. All the images are resized to $32\times 100$. With an NVIDIA GTX-1080Ti GPU, the MORAN takes 10.4ms to recognize an image containing five characters in lexicon-free mode.

\subsection{Performance of the MORAN}

We used a max-pooling layer at the top of the MORN. To evaluate the effectiveness of this technique, a comparison of pooling layers with different configurations is shown in Table \ref{table:comparison-of-pooling}. The accuracy is the highest when we use a max-pooling layer with a kernel size of 2 and stride of 1.

Before conducting a comparison with other methods, we list three results with a progressive combination of methods in Table \ref{table:Performance-of-MORAN}. The MORAN trained in an end-to-end manner already achieves very promising performance. In  curriculum learning, the first experiment is carried out using only an ASRN. Then, a MORN is added to the bottom of the above network to rectify the images. The last result is from the entire MORAN, including the MORN and ASRN trained with the fractional pickup method. The contribution of each part of our method is hence clearly demonstrated. For ICDAR OCR tasks, we report the total edit distance in Table \ref{table:Performance-of-MORAN(TED)}.

\begin{table}[h]
\centering
\caption{Performance of the MORAN (total edit distance). }
\label{table:Performance-of-MORAN(TED)}
\begin{tabular}{|c|c|c|c|}
\hline
\multirow{1}{*}{Method} & IC03 & IC13 & IC15 \\
\cline{2-4}
\hline
End-to-end training & 29.1 & 57.7 & 368.8 \\
\hline
Only ASRN & 33.8 & 69.1 & 376.8 \\
MORAN without FP & 22.7 & 45.3 & 345.2 \\
MORAN with FP & \textbf{19.8} & \textbf{42.0} & \textbf{334.0} \\
\hline
\end{tabular}
\end{table}

\subsection{Comparisons with Rectification Methods}
\textbf{Affine Transformation}: The results using the affine transformation are provided by Liu et al. \cite{liu2016star}. For fair comparison, we replace the ASRN by the R-Net proposed by Liu et al. \cite{liu2016star}. A direct comparison of the results is shown in Table \ref{table:comparison-stn}. As demonstrated in Fig.\ref{fig:compare-stn} and described in Section  \ref{section:Multi-Object Rectification Network}, affine transformation is limited by the geometric constraints of rotation, scaling and translation. However, the distortion of scene text is complicated. The MORAN is more flexible than affine transformation. It is able to predict smooth rectification for images free of geometric constraints.

\begin{table}[h]
\centering
\caption{Comparison with STAR-Net. }
\label{table:comparison-stn}
\begin{tabular}{|p{2.2cm}<{\centering}|p{0.9cm}<{\centering}|p{0.6cm}<{\centering}|p{0.6cm}<{\centering}| p{0.6cm}<{\centering}|p{1.1cm}<{\centering}|}
\hline
\multirow{1}{*}{Method} & IIIT5K & SVT & IC03 & IC13 & SVT-P \\
\cline{2-6}
\hline
Liu et al. \cite{liu2016star} & 83.3 & 83.6 & 89.9 & \textbf{89.1} & 73.5 \\
\hline
Ours & \textbf{87.5} & \textbf{83.9} & \textbf{92.5} & \textbf{89.1} & \textbf{74.6} \\
\hline
\end{tabular}
\end{table}

\textbf{RARE} \cite{shi2016robust}: The results of RARE given by Shi et al. \cite{shi2016robust} are in the Table \ref{table:Results on general benchmarks} and Table \ref{table:Results on irregular text}. We directly compare the network using exactly the same recognition network as that proposed in RARE. The results are shown in Table \ref{table:comparison-rare}.

\begin{figure}[h]
\centering
\begin{overpic}[width=8.5cm,height=5cm]{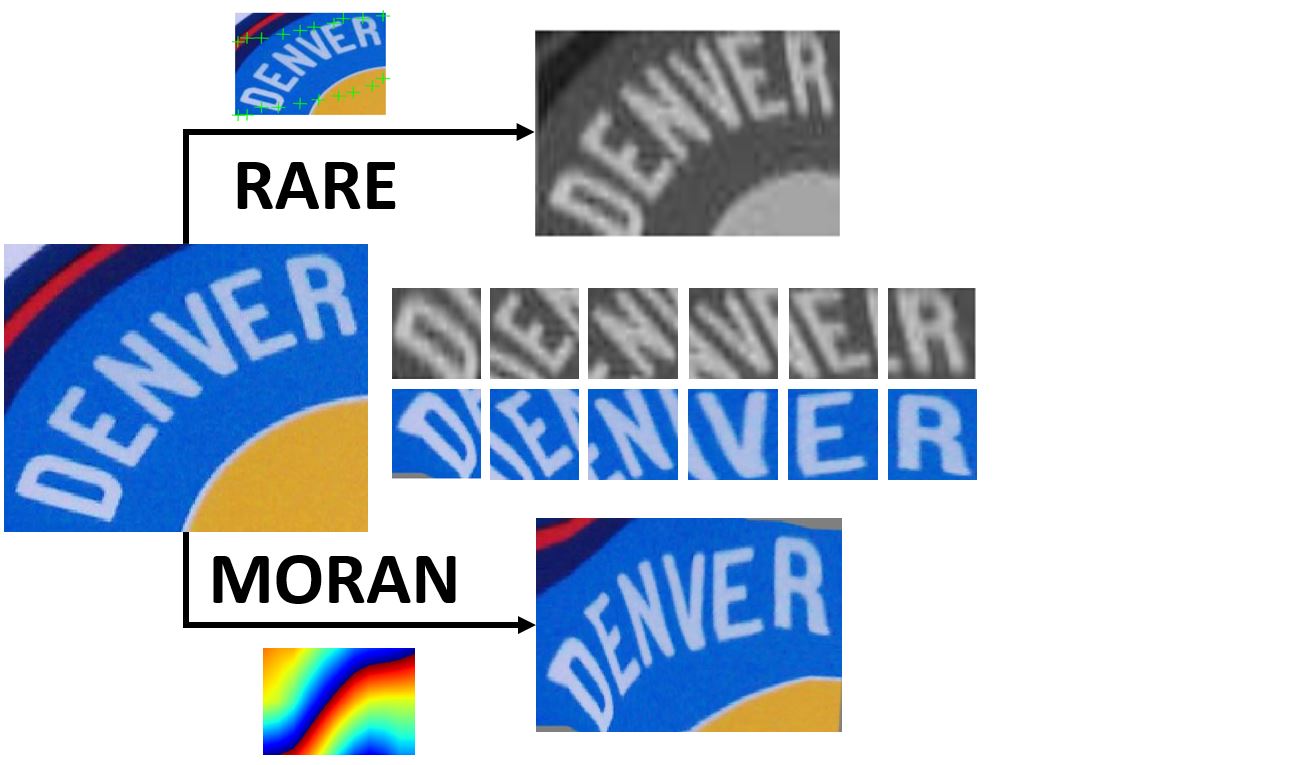}

\put(70,50){Predict:\color{red}{stink}\color{green}{er}}
\put(75,30){GT:denver}
\put(70,10){Predict:\color{green}{denver}}

\end{overpic}

\caption{Comparison of the MORAN and RARE. All characters are cropped for further comparison. The recognition results are on the right. ``GT" denotes the ground truth.}

\label{fig:comparison-other}
\end{figure}

The MORAN has some benefits and drawbacks comparing with RARE. RARE using fiducial points can only capture the overall text shape of an input image, whereas the MORAN can rectify every character in an image. As shown in Fig. \ref{fig:comparison-other}, all the characters in the image rectified by the MORAN are more normal in appearance than those of RARE. Furthermore, the MORAN without any fiducial points is theoretically able to rectify text of infinite length.

\begin{table*}[t]
\centering
\caption{Comparison with RARE. }
\label{table:comparison-rare}
\begin{tabular}{|c|c|c|c|c|c|c|}
\hline
\multirow{1}{*}{Method} & IIIT5K & SVT & IC03 & IC13 & SVT-P & CUTE80 \\
\cline{2-7}
\hline
Shi et al. \cite{shi2016robust} & 81.9 & 81.9 & 90.1 & 88.6 & 71.8 & 59.2 \\
\hline
Ours & \textbf{87.9} & \textbf{83.9} & \textbf{92.7} & \textbf{90.0} & \textbf{73.2} & \textbf{72.6} \\
\hline
\end{tabular}
\end{table*}

The training of MORAN is more difficult than that of RARE. We thus designed a curriculum learning strategy to enable the stable convergence of the MORAN. In terms of RARE, although it is end-to-end optimized with special initialization, randomly initialized network may result in failure of convergence.

\begin{table*}[t]
\centering
\caption{Results on general benchmarks. ``50" and ``1k" are lexicon sizes. ``Full" indicates the combined lexicon of all images in the benchmarks. ``None" means lexicon-free.}
\label{table:Results on general benchmarks}
\begin{tabular}{| c | c c c | c c | c c c | c |}
\hline
\multirow{2}{*}{Method} & \multicolumn{3}{c|}{IIIT5K} & \multicolumn{2}{c|}{SVT} &\multicolumn{3}{c|}{IC03} & IC13 \\
\cline{2-10}
 & 50 & 1k & None & 50 & None & 50 & Full & None & None \\
\hline
Almaz\'{a}n et al \cite{almazan2014word} & 91.2 & 82.1 & - & 89.2 & - & - & - & - & - \\
Yao et al. \cite{yao2014strokelets} & 80.2 & 69.3 & - & 75.9 & - & 88.5 & 80.3 & - & - \\
R.-Serrano et al. \cite{rodriguez2015label} & 76.1 & 57.4 & - & 70.0 & - & - & - & - & - \\
Jaderberg et al. \cite{jaderberg2014deep} & - & - & - & 86.1 & - & 96.2 & 91.5 & - & - \\
Su and Lu \cite{su2014accurate} & - & - & - & 83.0 & - & 92.0 & 82.0 & - & - \\
Gordo \cite{gordo2015supervised} & 93.3 & 86.6 & - & 91.8 & - & - & - & - & -  \\
Jaderberg et al. \cite{Jaderberg2015Deep} & 95.5 & 89.6 & - & 93.2 & 71.7 & 97.8 & 97.0 & 89.6 & 81.8 \\
Jaderberg et al. \cite{jaderberg2016reading} & 97.1 & 92.7 & - & 95.4 & 80.7* & \textbf{98.7} & \textbf{98.6} & 93.1* & 90.8* \\
Shi, Bai, and Yao \cite{shi2017end} & 97.8 & 95.0 & 81.2 & \textbf{97.5} & 82.7 & \textbf{98.7} & 98.0 & 91.9 & 89.6 \\
Shi et al. \cite{shi2016robust} & 96.2 & 93.8 & 81.9 & 95.5 & 81.9 & 98.3 & 96.2 & 90.1 & 88.6 \\
Lee and Osindero \cite{lee2016recursive} & 96.8 & 94.4 & 78.4 & 96.3 & 80.7 & 97.9 & 97.0 & 88.7 & 90.0 \\
Liu et al. \cite{liu2016star} & 97.7 & 94.5 & 83.3 & 95.5 & 83.6 & 96.9 & 95.3 & 89.9 & 89.1 \\
Yang et al. \cite{yang2017learning} & 97.8 & 96.1 & - & 95.2 & - & 97.7 & - & - & -\\
Yin et al. \cite{yin2017scene} & 98.7 & 96.1 & 78.2 & 95.1 & 72.5 & 97.6 & 96.5 & 81.1 & 81.4 \\
Cheng et al. \cite{cheng2017focusing} & 98.9 & 96.8 & 83.7 & 95.7 & 82.2 & 98.5 & 96.7 & 91.5 & 89.4 \\
Cheng et al. \cite{cheng2017arbitrarily} & \textbf{99.6} & \textbf{98.1} & 87.0 & 96.0 & 82.8 & 98.5 & 97.1 & 91.5 & - \\
\hline
Ours & 97.9 & 96.2 & \textbf{91.2} & 96.6 & \textbf{88.3} & \textbf{98.7} & 97.8 & \textbf{95.0} & \textbf{92.4} \\
\hline
\end{tabular}
\end{table*}

\subsection{Results on General Benchmarks}
The MORAN was evaluated on general benchmarks in which most of the testing samples are regular text and a small part of them are irregular text. The MORAN was compared with 16 methods and the results are shown in Table \ref{table:Results on general benchmarks}.

In Table \ref{table:Results on general benchmarks}, the MORAN outperforms all current state-of-the-art methods in lexicon-free mode. As Jaderberg \cite{jaderberg2016reading} treated each word as a category and the model cannot predict out-of-vocabulary words, we highlight these results by adding an asterisk. FAN \cite{cheng2017focusing} trained with pixel-level supervision is also beyond the scope of consideration. We hence compare the MORAN with the baseline of FAN.

\begin{table*}[t]
\centering
\caption{Results on irregular datasets. ``50" is lexicon sizes. ``Full" indicates the combined lexicon of all images in the benchmarks. ``None" means lexicon-free.}
\label{table:Results on irregular text}
\begin{tabular}{|c|p{1.cm}<{\centering} p{1.cm}<{\centering} p{1.cm}<{\centering} |p{1.5cm}<{\centering}|p{1.5cm}<{\centering}|}
\hline
\multirow{2}{*}{Method} & \multicolumn{3}{c|}{SVT-Perspective} & CUTE80 & IC15 \\
\cline{2-6}
 & 50 & Full & None & None & None \\
\hline
ABBYY et al. \cite{wang2011end} & 40.5 & 26.1 & - & - & - \\
Mishra et al. \cite{mishra2012scene} & 45.7 & 24.7 & - & - & - \\
Wang et al. \cite{wang2012end} & 40.2 & 32.4 & - & - & - \\
Phan et al. \cite{quy2013recognizing} & 75.6 & 67.0 & - & - & - \\
Shi et al. \cite{shi2016robust} & 91.2 & 77.4 & 71.8 & 59.2 & - \\
Yang et al. \cite{yang2017learning} & 93.0 & 80.2 & 75.8 & 69.3 & - \\
Liu et al. \cite{liu2016star} & \textbf{94.3} & 83.6 & 73.5 & - & - \\
Cheng et al. \cite{cheng2017focusing} & 92.6 & 81.6 & 71.5 & 63.9 & 66.2 \\
Cheng et al. \cite{cheng2017arbitrarily} & 94.0 & 83.7 & 73.0 & 76.8 & 68.2 \\
\hline
Ours & \textbf{94.3} & \textbf{86.7} & \textbf{76.1} & \textbf{77.4} & \textbf{68.8}\\
\hline
\end{tabular}
\end{table*}
\subsection{Results on Irregular Text}

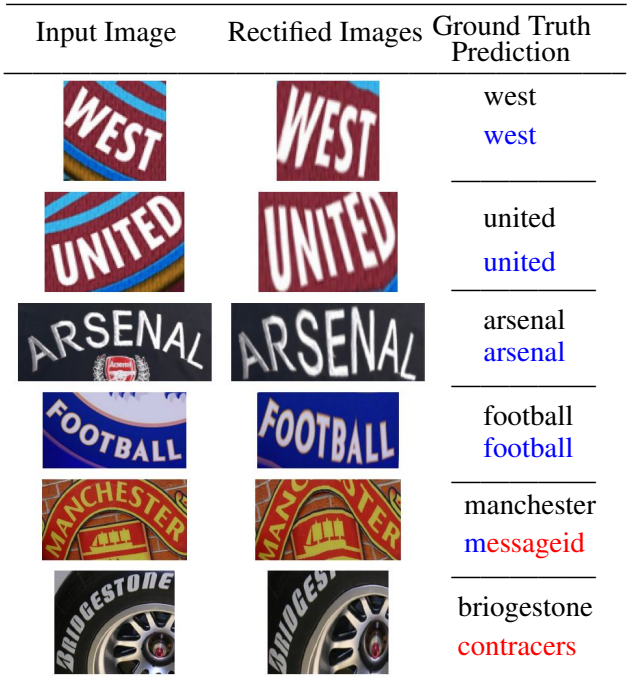
\begin{figure}[t]
\rule{8.25cm}{0.05em}
\\
\\
\\
\centering
\begin{overpic}[width=8.5cm,height=8cm]{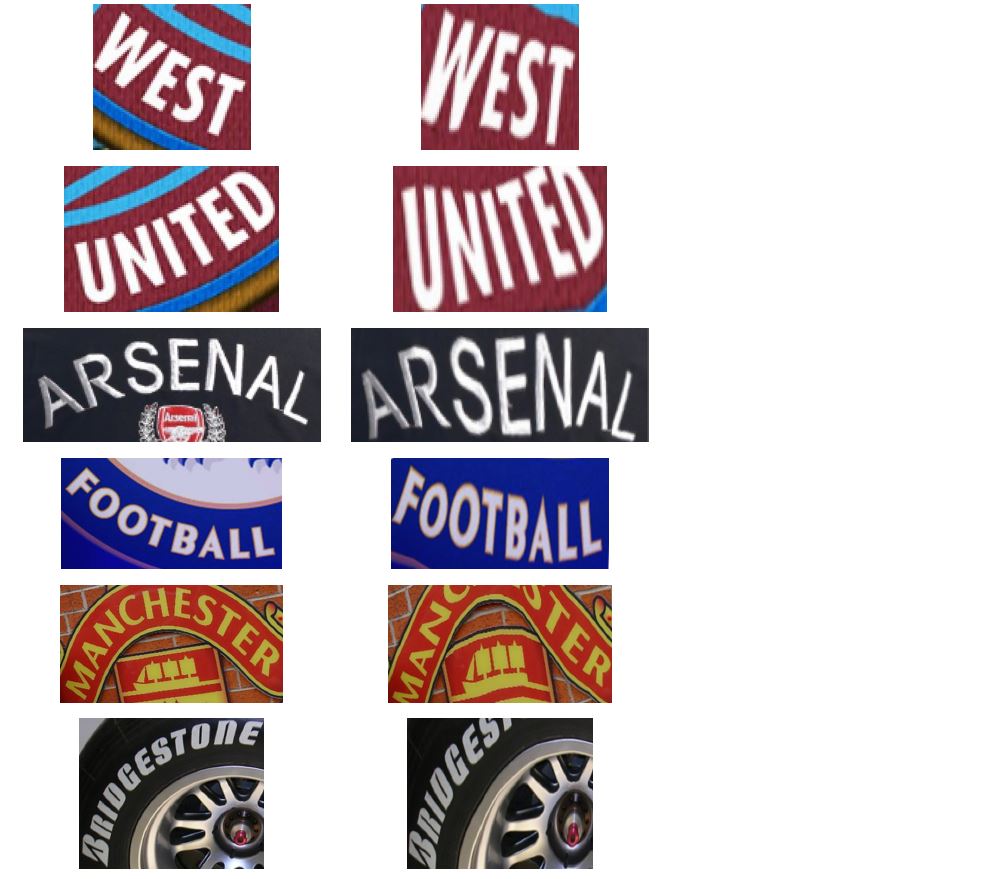}

\put(5,100){Input Image}
\put(35,100){Rectified Images}
\put(67,101){Ground Truth}
\put(70,97){Prediction}
\put(0,94){-----------------------------------------------------------------}

\put(75,90){west}
\put(75,84){\color{blue}{west}}
\put(70,77){---------------}

\put(75,71){united}
\put(75,64){\color{blue}{united}}
\put(70,60){---------------}

\put(75,55){arsenal}
\put(75,50){\color{blue}{arsenal}}
\put(70,45){---------------}

\put(75,40){football}
\put(75,35){\color{blue}{football}}
\put(70,30){---------------}

\put(72,26){manchester}
\put(72,20){\color{blue}{m}\color{red}{essageid}}
\put(70,15){---------------}

\put(71,10){briogestone}
\put(71,4){\color{red}{contracers}}

\put(0,-2){-----------------------------------------------------------------}
\end{overpic}

\caption{Effects of different curve angles of scene text. The first four rows are text with small curve angles and the last two rows are text with large curve angles. The MORAN can rectify irregular text with small curve angles.}

\label{fig:6-irregular-samples}
\end{figure}

The MORAN was also evaluated on irregular text datasets to reveal the contribution of the MORN. The results on SVT-Perspective, CUTE80 and IC15 are shown in Table \ref{table:Results on irregular text}. The MORAN is still the best of all methods.

For the SVT-Perspective dataset, many samples are low-resolution and perspective. The result of the MORAN with 50-word lexicon is the same as that of the method of Liu et al. \cite{liu2016star}. However, the MORAN outperforms all methods in the setting without any lexicon.

In addition to perspective text, the MORAN is able to recognize curved text. Some examples are demonstrated in Fig. \ref{fig:6-irregular-samples}. The MORAN is able to rectify most curved text in CUTE80 and correctly recognize them. It is hence adequately robust to rectify text with small curve angle.

\subsection{Limitation of the MORAN}

For fair comparisons and good repeatability, we chose the widely used training datasets, which contain only horizontal synthetic text. Therefore, because of complicated background, the MORAN will fail when the curve angle is too large. Such cases are given in the last two rows of Fig. \ref{fig:6-irregular-samples}. MORAN mistakenly regards the complicated background as foreground. However, such samples are rare in training datasets.

Furthermore, with the existing training datasets and without any data augmentation, the MORAN focuses more on horizontal irregular text. Note that there are many vertical text in IC15. However, the MORAN is not designed for vertical text. Our method was proposed for the complicated deformation of text within a cropped horizontal rectangle.

The experiments above are all based on cropped text recognition. A MORAN without a text detector is not an end-to-end scene text recognition system. Actually, in more application scenarios, irregular and multi-oriented text are challenging both for detection and recognition, which have attracted great interest. For instance, Liu et al. \cite{yuliang2017detecting} and Ch'ng et al. \cite{CK2017} released complicated datasets. Sain et al. \cite{sain2018multi} and He et al. \cite{he2018multi} proposed methods to improve the performance of multi-oriented text detection. Therefore, scene text recognition still remains a challenging problem waiting for solutions.

\section{Conclusion}
In this paper, we presented a multi-object rectified attention network (MORAN) for scene text recognition. The proposed framework involves two stages: rectification and recognition. First, a multi-object rectification network, which is free of geometric constraints and flexible enough to handle complicated deformations, was proposed to transform an image containing irregular text into a more readable one. The rectified patterns decrease the difficulty of recognition. Then, an attention-based sequence recognition network was designed to recognize the rectified image and outputs the characters in sequence. Moreover, a fractional pickup method was proposed to expand the visual field of the attention-based decoder. The attention-based decoder thus obtains more context information and gains robustness. To efficiently train the network, we designed a curriculum learning strategy to respectively strengthen each sub-network. The proposed MORAN is trained in a weak-supervised way, which requires only images and the corresponding text labels. Experiments on both regular and irregular datasets, including IIIT5K, SVT, ICDAR2003, ICDAR2013, ICDAR2015, SVT-Perspective and CUTE80, demonstrate the outstanding performance of the MORAN.

In future, it is worth extending this method to deal with arbitrary-oriented text recognition, which is more challenging due to the wide variety of text and background. Moreover, the improvements in end-to-end text recognition performance come not just from the recognition model, but also from detection model. Therefore, finding a proper and effective way to combine the MORAN with a scene text detector is also a direction worth of study.

\section*{Acknowledgement}
This research was supported by the National Key R\&D Program of China (Grant No.: 2016YFB1001405), GD-NSF (Grant No.: 2017A030312006), NSFC (Grant No.: 61472144, 61673182), GDSTP (Grant No.: 2015B010101004, 2015B010130003, 2017A030312006), GZSTP (Grant No.: 201607010227).


{\small
\bibliographystyle{ieee}
\bibliography{mybibfile}

\begin{thebibliography}{10}\itemsep=-1pt

\bibitem{almazan2014word}
J.~Almaz{\'a}n, A.~Gordo, A.~Forn{\'e}s, and E.~Valveny.
\newblock Word spotting and recognition with embedded attributes.
\newblock {\em IEEE Trans. Pattern Anal. Mach. Intell.}, 36(12):2552--2566,
  2014.

\bibitem{bahdanau2014neural}
D.~Bahdanau, K.~Cho, and Y.~Bengio.
\newblock Neural machine translation by jointly learning to align and
  translate.
\newblock {\em CoRR}, abs/1409.0473 (2014).

\bibitem{bissacco2013photoocr}
A.~Bissacco, M.~Cummins, Y.~Netzer, and H.~Neven.
\newblock Photoocr: Reading text in uncontrolled conditions.
\newblock In {\em Proceedings of International Conference on Computer Vision
  (ICCV)}, pages 785--792, 2013.

\bibitem{bookstein1989principal}
F.~L. Bookstein.
\newblock Principal warps: Thin-plate splines and the decomposition of
  deformations.
\newblock {\em IEEE Trans. Pattern Anal. Mach. Intell.}, 11(6):567--585, 1989.

\bibitem{cheng2017focusing}
Z.~Cheng, F.~Bai, Y.~Xu, G.~Zheng, S.~Pu, and S.~Zhou.
\newblock Focusing attention: Towards accurate text recognition in natural
  images.
\newblock In {\em Proceedings of International Conference on Computer Vision
  (ICCV)}, pages 5086--5094, 2017.

\bibitem{cheng2017arbitrarily}
Z.~Cheng, Y.~Xu, F.~Bai, Y.~Niu, S.~Pu, and S.~Zhou.
\newblock {AON}: Towards arbitrarily-oriented text recognition.
\newblock In {\em Proceedings of the IEEE Conference on Computer Vision and
  Pattern Recognition (CVPR)}, pages 5571--5579, 2018.

\bibitem{CK2017}
C.~K. Ch'ng and C.~S. Chan.
\newblock Total-text: A comprehensive dataset for scene text detection and
  recognition.
\newblock In {\em Proceedings of International Conference on Document Analysis
  and Recognition (ICDAR)}, pages 935--942, 2017.

\bibitem{cho2014learning}
K.~Cho, B.~van Merrienboer, {\c{C}}.~G{\"{u}}l{\c{c}}ehre, D.~Bahdanau,
  F.~Bougares, H.~Schwenk, and Y.~Bengio.
\newblock Learning phrase representations using {RNN} encoder-decoder for
  statistical machine translation.
\newblock In {\em Proceedings of the 2014 Conference on Empirical Methods in
  Natural Language Processing, ({EMNLP})}, pages 1724--1734, 2014.

\bibitem{deformable2017}
J.~Dai, H.~Qi, Y.~Xiong, Y.~Li, G.~Zhang, H.~Hu, and Y.~Wei.
\newblock Deformable convolutional networks.
\newblock In {\em Proceedings of International Conference on Computer Vision
  (ICCV)}, pages 764--773, 2017.

\bibitem{dalal2005histograms}
N.~Dalal and B.~Triggs.
\newblock Histograms of oriented gradients for human detection.
\newblock In {\em Proceedings of Computer Vision and Pattern Recognition
  (CVPR)}, pages 886--893, 2005.

\bibitem{gomez2017textproposals}
L.~G{\'o}mez and D.~Karatzas.
\newblock Textproposals: a text-specific selective search algorithm for word
  spotting in the wild.
\newblock {\em Pattern Recognit.}, 70:60--74, 2017.

\bibitem{gordo2015supervised}
A.~Gordo.
\newblock Supervised mid-level features for word image representation.
\newblock In {\em Proceedings of Computer Vision and Pattern Recognition
  (CVPR)}, pages 2956--2964, 2015.

\bibitem{graves2006connectionist}
A.~Graves, S.~Fern{\'a}ndez, F.~Gomez, and J.~Schmidhuber.
\newblock Connectionist temporal classification: labelling unsegmented sequence
  data with recurrent neural networks.
\newblock In {\em Proceedings of International Conference on Machine Learning
  (ICML)}, pages 369--376, 2006.

\bibitem{gu2017recent}
J.~Gu, Z.~Wang, J.~Kuen, L.~Ma, A.~Shahroudy, B.~Shuai, T.~Liu, X.~Wang,
  G.~Wang, J.~Cai, et~al.
\newblock Recent advances in convolutional neural networks.
\newblock {\em Pattern Recognit.}, 77:354--377, 2018.

\bibitem{gupta2016synthetic}
A.~Gupta, A.~Vedaldi, and A.~Zisserman.
\newblock Synthetic data for text localisation in natural images.
\newblock In {\em Proceedings of Computer Vision and Pattern Recognition
  (CVPR)}, pages 2315--2324, 2016.

\bibitem{he2016deep}
K.~He, X.~Zhang, S.~Ren, and J.~Sun.
\newblock Deep residual learning for image recognition.
\newblock In {\em Proceedings of Computer Vision and Pattern Recognition
  (CVPR)}, pages 770--778, 2016.

\bibitem{he2016reading}
P.~He, W.~Huang, Y.~Qiao, C.~C. Loy, and X.~Tang.
\newblock Reading scene text in deep convolutional sequences.
\newblock In {\em Proceedings of Association for the Advancement of Artificial
  Intelligence (AAAI)}, pages 3501--3508, 2016.

\bibitem{he2018multi}
W.~He, X.-Y. Zhang, F.~Yin, and C.-L. Liu.
\newblock Multi-oriented and multi-lingual scene text detection with direct
  regression.
\newblock {\em IEEE Trans. Image Processing}, 27(11):5406--5419, 2018.

\bibitem{hochreiter1997long}
S.~Hochreiter and J.~Schmidhuber.
\newblock Long short-term memory.
\newblock {\em Neural computation}, 9(8):1735--1780, 1997.

\bibitem{jaderberg2014synthetic}
M.~Jaderberg, K.~Simonyan, A.~Vedaldi, and A.~Zisserman.
\newblock Synthetic data and artificial neural networks for natural scene text
  recognition.
\newblock In {\em Proceedings of Advances in Neural Information Processing Deep
  Learn. Workshop (NIPS-W)}, 2014.

\bibitem{Jaderberg2015Deep}
M.~Jaderberg, K.~Simonyan, A.~Vedaldi, and A.~Zisserman.
\newblock Deep structured output learning for unconstrained text recognition.
\newblock In {\em Proceedings of International Conference on Learning
  Representations (ICLR)}, 2015.

\bibitem{jaderberg2016reading}
M.~Jaderberg, K.~Simonyan, A.~Vedaldi, and A.~Zisserman.
\newblock Reading text in the wild with convolutional neural networks.
\newblock {\em International Journal of Computer Vision}, 116(1):1--20, 2016.

\bibitem{jaderberg2014deep}
M.~Jaderberg, A.~Vedaldi, and A.~Zisserman.
\newblock Deep features for text spotting.
\newblock In {\em Proceedings of European Conference on Computer Vision
  (ECCV)}, pages 512--528, 2014.

\bibitem{karatzas2015icdar}
D.~Karatzas, L.~Gomez-Bigorda, A.~Nicolaou, S.~Ghosh, A.~Bagdanov, M.~Iwamura,
  J.~Matas, L.~Neumann, V.~R. Chandrasekhar, S.~Lu, et~al.
\newblock {ICDAR} 2015 competition on robust reading.
\newblock In {\em Proceedings of International Conference on Document Analysis
  and Recognition (ICDAR)}, pages 1156--1160, 2015.

\bibitem{karatzas2013icdar}
D.~Karatzas, F.~Shafait, S.~Uchida, M.~Iwamura, L.~G. i~Bigorda, S.~R. Mestre,
  J.~Mas, D.~F. Mota, J.~A. Almazan, and L.~P. De~Las~Heras.
\newblock {ICDAR} 2013 robust reading competition.
\newblock In {\em Proceedings of International Conference on Document Analysis
  and Recognition (ICDAR)}, pages 1484--1493, 2013.

\bibitem{khare2016blind}
V.~Khare, P.~Shivakumara, P.~Raveendran, and M.~Blumenstein.
\newblock A blind deconvolution model for scene text detection and recognition
  in video.
\newblock {\em Pattern Recognit.}, 54:128--148, 2016.

\bibitem{lee2016recursive}
C.-Y. Lee and S.~Osindero.
\newblock Recursive recurrent nets with attention modeling for {OCR} in the
  wild.
\newblock In {\em Proceedings of Computer Vision and Pattern Recognition
  (CVPR)}, pages 2231--2239, 2016.

\bibitem{liu2016star}
W.~Liu, C.~Chen, K.-Y.~K. Wong, Z.~Su, and J.~Han.
\newblock {STAR-Net}: A spatial attention residue network for scene text
  recognition.
\newblock In {\em Proceedings of British Machine Vision Conference (BMVC)},
  page~7, 2016.

\bibitem{yuliang2017detecting}
Y.~Liu, L.~Jin, S.~Zhang, and S.~Zhang.
\newblock Detecting curve text in the wild: New dataset and new solution.
\newblock {\em CoRR}, abs/1712.02170 (2017), 2017.

\bibitem{lou2016generative}
X.~Lou, K.~Kansky, W.~Lehrach, C.~Laan, B.~Marthi, D.~Phoenix, and D.~George.
\newblock Generative shape models: Joint text recognition and segmentation with
  very little training data.
\newblock In {\em Proceedings of Advances in Neural Information Processing
  Systems (NIPS)}, pages 2793--2801, 2016.

\bibitem{lucas2003icdar}
S.~M. Lucas, A.~Panaretos, L.~Sosa, A.~Tang, S.~Wong, and R.~Young.
\newblock {ICDAR} 2003 robust reading competitions.
\newblock In {\em Proceedings of International Conference on Document Analysis
  and Recognition (ICDAR)}, pages 682--687, 2003.

\bibitem{mishra2012scene}
A.~Mishra, K.~Alahari, and C.~Jawahar.
\newblock Scene text recognition using higher order language priors.
\newblock In {\em Proceedings of British Machine Vision Conference (BMVC)},
  pages 1--11, 2012.

\bibitem{neumann2012real}
L.~Neumann and J.~Matas.
\newblock Real-time scene text localization and recognition.
\newblock In {\em Proceedings of Computer Vision and Pattern Recognition
  (CVPR)}, pages 3538--3545, 2012.

\bibitem{pytorch}
A.~Paszke, S.~Gross, S.~Chintala, G.~Chanan, E.~Yang, Z.~DeVito, Z.~Lin,
  A.~Desmaison, L.~Antiga, and A.~Lerer.
\newblock Automatic differentiation in pytorch.
\newblock In {\em Proceedings of Advances in Neural Information Processing
  Systems Autodiff Workshop (NIPS-W)}, 2017.

\bibitem{quy2013recognizing}
T.~Quy~Phan, P.~Shivakumara, S.~Tian, and C.~Lim~Tan.
\newblock Recognizing text with perspective distortion in natural scenes.
\newblock In {\em Proceedings of International Conference on Computer Vision
  (ICCV)}, pages 569--576, 2013.

\bibitem{rabin2011wasserstein}
J.~Rabin, G.~Peyr{\'e}, J.~Delon, and M.~Bernot.
\newblock Wasserstein barycenter and its application to texture mixing.
\newblock In {\em Proceedings of International Conference on Scale Space and
  Variational Methods (ICSSVM)}, pages 435--446, 2011.

\bibitem{risnumawan2014robust}
A.~Risnumawan, P.~Shivakumara, C.~S. Chan, and C.~L. Tan.
\newblock A robust arbitrary text detection system for natural scene images.
\newblock {\em Expert Systems with Applications}, 41(18):8027--8048, 2014.

\bibitem{rodriguez2015label}
J.~A. Rodriguez-Serrano, A.~Gordo, and F.~Perronnin.
\newblock Label embedding: A frugal baseline for text recognition.
\newblock {\em International Journal of Computer Vision}, 113(3):193--207,
  2015.

\bibitem{sain2018multi}
A.~Sain, A.~K. Bhunia, P.~P. Roy, and U.~Pal.
\newblock Multi-oriented text detection and verification in video frames and
  scene images.
\newblock {\em Neurocomputing}, 275:1531--1549, 2018.

\bibitem{seok2015scene}
J.-H. Seok and J.~H. Kim.
\newblock Scene text recognition using a hough forest implicit shape model and
  semi-markov conditional random fields.
\newblock {\em Pattern Recognit.}, 48(11):3584--3599, 2015.

\bibitem{shi2017end}
B.~Shi, X.~Bai, and C.~Yao.
\newblock An end-to-end trainable neural network for image-based sequence
  recognition and its application to scene text recognition.
\newblock {\em IEEE Trans. Pattern Anal. Mach. Intell.}, 39(11):2298--2304,
  2017.

\bibitem{shi2016robust}
B.~Shi, X.~Wang, P.~Lyu, C.~Yao, and X.~Bai.
\newblock Robust scene text recognition with automatic rectification.
\newblock In {\em Proceedings of Computer Vision and Pattern Recognition
  (CVPR)}, pages 4168--4176, 2016.

\bibitem{shi2014end}
C.~Shi, C.~Wang, B.~Xiao, S.~Gao, and J.~Hu.
\newblock End-to-end scene text recognition using tree-structured models.
\newblock {\em Pattern Recognit.}, 47(9):2853--2866, 2014.

\bibitem{su2014accurate}
B.~Su and S.~Lu.
\newblock Accurate scene text recognition based on recurrent neural network.
\newblock In {\em Proceedings of Asian Conference on Computer Vision (ACCV)},
  pages 35--48, 2014.

\bibitem{su2017accurate}
B.~Su and S.~Lu.
\newblock Accurate recognition of words in scenes without character
  segmentation using recurrent neural network.
\newblock {\em Pattern Recognit.}, 63:397--405, 2017.

\bibitem{sun2015robust}
L.~Sun, Q.~Huo, W.~Jia, and K.~Chen.
\newblock A robust approach for text detection from natural scene images.
\newblock {\em Pattern Recognit.}, 48(9):2906--2920, 2015.

\bibitem{sutskever2014sequence}
I.~Sutskever, O.~Vinyals, and Q.~V. Le.
\newblock Sequence to sequence learning with neural networks.
\newblock In {\em Proceedings of Advances in Neural Information Processing
  Systems (NIPS)}, pages 3104--3112, 2014.

\bibitem{wang2011end}
K.~Wang, B.~Babenko, and S.~Belongie.
\newblock End-to-end scene text recognition.
\newblock In {\em Proceedings of International Conference on Computer Vision
  (ICCV)}, pages 1457--1464, 2011.

\bibitem{wang2010word}
K.~Wang and S.~Belongie.
\newblock Word spotting in the wild.
\newblock In {\em Proceedings of European Conference on Computer Vision
  (ECCV)}, pages 591--604, 2010.

\bibitem{wang2012end}
T.~Wang, D.~J. Wu, A.~Coates, and A.~Y. Ng.
\newblock End-to-end text recognition with convolutional neural networks.
\newblock In {\em Proceedings of International Conference on Pattern
  Recognition (ICPR)}, pages 3304--3308, 2012.

\bibitem{yang2017learning}
X.~Yang, D.~He, Z.~Zhou, D.~Kifer, and C.~L. Giles.
\newblock Learning to read irregular text with attention mechanisms.
\newblock In {\em Proceedings of International Joint Conference on Artificial
  Intelligence, (IJCAI)}, pages 3280--3286, 2017.

\bibitem{yao2014strokelets}
C.~Yao, X.~Bai, B.~Shi, and W.~Liu.
\newblock Strokelets: A learned multi-scale representation for scene text
  recognition.
\newblock In {\em Proceedings of Computer Vision and Pattern Recognition
  (CVPR)}, pages 4042--4049, 2014.

\bibitem{ye2015text}
Q.~Ye and D.~Doermann.
\newblock Text detection and recognition in imagery: A survey.
\newblock {\em IEEE Trans. Pattern Anal. Mach. Intell.}, 37(7):1480--1500,
  2015.

\bibitem{yin2017scene}
F.~Yin, Y.~Wu, X.~Zhang, and C.~Liu.
\newblock Scene text recognition with sliding convolutional character models.
\newblock {\em CoRR}, abs/1709.01727 (2017), 2017.

\bibitem{zeiler2012adadelta}
M.~D. Zeiler.
\newblock {ADADELTA:} an adaptive learning rate method.
\newblock {\em CoRR}, abs/1212.5701 (2012), 2012.

\bibitem{zhu2016could}
A.~Zhu, R.~Gao, and S.~Uchida.
\newblock Could scene context be beneficial for scene text detection?
\newblock {\em Pattern Recognit.}, 58:204--215, 2016.

\bibitem{zhu2016scene}
Y.~Zhu, C.~Yao, and X.~Bai.
\newblock Scene text detection and recognition: Recent advances and future
  trends.
\newblock {\em Frontiers of Computer Science}, 10(1):19--36, 2016.

\end{thebibliography}
}

\end{document}